\documentclass[10pt,journal,compsoc]{IEEEtran}
\usepackage{amsmath,amsfonts}
\usepackage{algorithmic}
\usepackage{algorithm}
\usepackage{array}
\usepackage{textcomp}
\usepackage{stfloats}
\usepackage{url}
\usepackage{enumitem}
\usepackage{verbatim}
\usepackage{graphicx}
\usepackage{cite}
\usepackage{makecell}
\usepackage{siunitx}
\usepackage{multicol}
\usepackage{multirow}
\usepackage{forloop}
\usepackage{xspace}
\usepackage{wrapfig}
\usepackage{tabularx}
\usepackage{arydshln}
\usepackage{calc}
\usepackage{cprotect}
\usepackage{tabularray}
\usepackage{subcaption}
\usepackage[table]{xcolor}
\usepackage{color, colortbl,xcolor}
\usepackage{CJK}
\usepackage{booktabs}
\usepackage{xfrac}  

\definecolor{LightRed}{RGB}{255, 199, 206}
\definecolor{TRed}{RGB}{196, 49, 25}
\definecolor{TBlue}{RGB}{8, 111, 189}

\definecolor{AudioRed}{RGB}{255, 0, 0}
\definecolor{AudioGreen}{RGB}{84, 130, 53}

\def\mathbi#1{\textbf{\em #1}} 
\newcommand{\defcal}[1]{\expandafter\newcommand\csname 
	c#1\endcsname{{\mathcal{#1}}}}
\newcommand{\defbb}[1]{\expandafter\newcommand\csname 
	b#1\endcsname{{\mathbb{#1}}}}
\newcommand{\defbf}[1]{\expandafter\newcommand\csname 
	bf#1\endcsname{{\mathbf{#1}}}}
\newcommand{\defbk}[1]{\expandafter\newcommand\csname 
	bk#1\endcsname{{\mathfrak{#1}}}}
\newcommand{\defbi}[1]{\expandafter\newcommand\csname 
	bi#1\endcsname{{\mathbi{#1}}}}
\newcounter{calBbCounter}
\forLoop{1}{26}{calBbCounter}{
	\edef\letter{\Alph{calBbCounter}}
	\expandafter\defcal\letter
	\expandafter\defbb\letter
	\expandafter\defbf\letter
	\expandafter\defbk\letter
	\expandafter\defbi\letter
}
\usepackage[pagebackref,breaklinks,colorlinks]{hyperref}

\usepackage[capitalize]{cleveref}
\crefname{section}{Sec.}{Secs.}
\Crefname{section}{Section}{Sections}
\Crefname{table}{Table}{Tables}
\crefname{table}{Tab.}{Tabs.}
\newcommand{\RomanNumeral}[1]{\romannumeral #1}
\newcommand{\RomanNumeralCaps}[1]{\uppercase\expandafter{\romannumeral#1}}

\makeatletter
\DeclareRobustCommand\onedot{\futurelet\@let@token\@onedot}
\def\@onedot{\ifx\@let@token.\else.\null\fi\xspace}

\def\eg{\emph{e.g}\onedot} 
\def\ie{\emph{i.e}\onedot} 
 
\def\etc{\emph{etc}\onedot}

\def\etal{\emph{et al}\onedot}

\newcommand{\FixedMM}[1][]{\cI^{#1}}
\newcommand{\DynamMM}[1][]{m^{#1}}

\newcommand{\AudioFeat}[1]{s_{#1}}

\newcommand{\SpeakerFeat}[1]{m_{#1}^s}

\newcommand{\ListenerFeat}[1]{m_{#1}^l}

\newcommand{\RoleFeat}[1]{m_{#1}}

\newcommand{\RoleFeatsT}{\cM_T}
\newcommand{\RoleFeatHat}[1]{\hat{m}_{#1}}

\newcommand{\RoleFeatsHatT}{\hat{\cM}_T}

\newcommand{\Generator}{\bfG_m}
\newcommand{\Render}{\bfG_v}

\definecolor{apositive}{RGB}{248, 203, 172}
\definecolor{anegative}{RGB}{180, 199, 231}
\definecolor{aneutral}{RGB}{197, 224, 181}

\usepackage{pifont}
\newcommand{\cmark}{\ding{51}}
\newcommand{\xmark}{\ding{55}}
\begin{document}

\title{Interactive Conversational Head Generation}

\author{Mohan~Zhou,
        ~Yalong~Bai,
        ~Wei Zhang,
        Ting Yao,~\IEEEmembership{Senior Member,~IEEE}
        and~Tiejun~Zhao% <-this % stops a space
\IEEEcompsocitemizethanks{
\IEEEcompsocthanksitem Corresponding Email: tjzhao@hit.edu.cn
}% <-this % stops an unwanted space
}

% \markboth{Journal of \LaTeX\ Class Files,~Vol.~14, No.~8, August~2015}%
% {Shell \MakeLowercase{\textit{et al.}}: Bare Demo of IEEEtran.cls for Computer Society Journals}
\markboth{}{}


\IEEEtitleabstractindextext{%
\begin{abstract}
We introduce a new conversation head generation benchmark for synthesizing behaviors of a single interlocutor in a face-to-face conversation. The capability to automatically synthesize interlocutors which can participate in long and multi-turn conversations is vital and offer benefits for various applications, including digital humans, virtual agents, and social robots. While existing research primarily focuses on talking head generation (one-way interaction), hindering the ability to create a digital human for conversation (two-way) interaction due to the absence of listening and interaction parts. 
In this work, we construct two datasets to address this issue, ``ViCo'' for independent talking and listening head generation tasks at the sentence level, and ``ViCo-X'', for synthesizing interlocutors in multi-turn conversational scenarios.
Based on ViCo and ViCo-X, we define three novel tasks targeting the interaction modeling during the face-to-face conversation: 1) responsive listening head generation making listeners respond actively to the speaker with non-verbal signals, 2) expressive talking head generation guiding speakers to be aware of listeners' behaviors, and 3) conversational head generation to integrate the talking/listening ability in one interlocutor. 
Along with the datasets, we also propose corresponding baseline solutions to the three aforementioned tasks.
Experimental results show that our baseline method could generate responsive and vivid agents that can collaborate with real person to fulfil the whole conversation.
Project page: \url{https://vico.solutions/}.
\end{abstract}
\begin{IEEEkeywords}
Conversational Head Generation, Listening Head Generation, Video Synthesis
\end{IEEEkeywords}}

\maketitle
\IEEEdisplaynontitleabstractindextext
\IEEEpeerreviewmaketitle

\IEEEraisesectionheading{\section{Introduction}}
\label{sec:intro}
\IEEEPARstart{C}{ommunication}~\cite{berger2005interpersonal,honeycutt2001mental,parker2000improving,stacks2014integrated,tomasello2010origins} is the fundamental social process for effectively exchanging information between people. A typical oral-based conversation ordinarily involves two interlocutors alternating the roles of speaker and listener to achieve a successful conversation through verbal and non-verbal reciprocal interactions, \eg, auditory signals, gestures, hand signs, body postures, and \etc. There have been extensive investigations into the study of human communication among sociology, psychology, human-computer interaction, and \etc. Especially the oral-based conversation can be modeled by three dimensions: \emph{speaker}, \emph{listener}, and their mutual \emph{interactions}. In real scenarios, all three dimensions are equally important: speakers produce and transmit messages, listeners receive messages and give feedback, and the mutual interactions connect them, making the conversation a closed loop. 

\begin{figure*}[t]
    \centering
    \includegraphics[width=\linewidth]{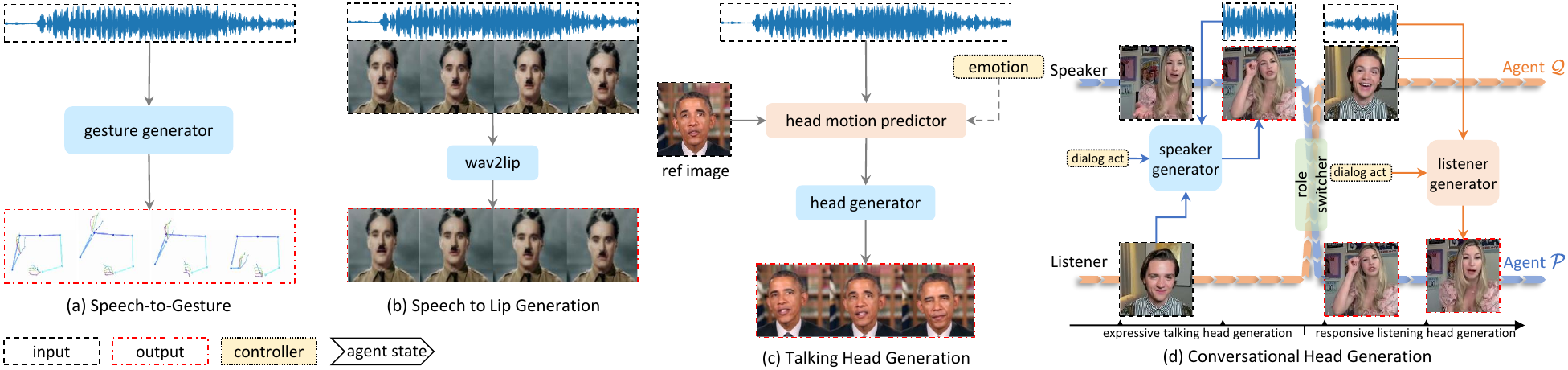}
    \caption{Illustration of three related tasks (a $\sim$ c) and our proposed responsive listening head generation and conversational head generation (d). (a) Speech-to-gesture translation: generates plausible body gestures to go along with the input speech. (b) Speech to lip generation: produces synchronized lip motions based on given speech. (c) Talking head generation: synthesizes talking face video conditioned on speaker identity, speech audio, and/or the speaker's emotion. (d) Our proposed conversational head generation involves both talking and listening head generation in mutual interaction, responding to the other interlocutor during a multi-turn conversation.}
    \label{fig:taskdescription}
\end{figure*}

While existing researches mainly focus on single-sided communication -- speaker-centric synthesis. As shown in \cref{fig:taskdescription}, speech-to-gesture generation~\cite{ginosar2019learning} learns a mapping between the audio signal and the speaker's pose. Speech to lip generation~\cite{prajwal2020lip} aims to refine the lip-synchronization of a given video input. Talking head synthesis~\cite{chung2017you,wang2020mead,zhu2021deep} tries to generate a vivid talking video of a specific speaker with facial animations from a still image and a clip of audio. However, these works solely concentrate on the speaking role, disregarding the essential role of the listener, let alone the crucial aspect of their mutual interactions. Notably, during a face-to-face conversation, listening behavior is even more critical, as proper feedback to the speaker (\eg, \textit{nod}, \textit{smile}, \textit{eye contact}, \etc.) is vital for a successful communication~\cite{mcnaughton2008learning,robertson2005active,rost2013active}. Through real-time feedback, listeners show how they are engaged (\eg, \textit{interested}, \textit{understand}, \textit{agree}, \etc.) to the speech, such that conversation gets more accessible for both participants. And only if both the listener and the speaker are modeled, do we have the occasion to model their interactions.

In this paper, we first construct two datasets for the conversational head generation: 1) \textbf{ViCo}, a dataset of in-the-wild speaker-listener clip pairs to highlight the listening part within one utterance. Three listening styles, based on positive, neutral, and negative attitudes, are exhibited by listeners, and qualified listeners are expected to respond actively to the speaker with verbal and non-verbal signals. It serves the purpose of talking and listening head generation at the sentence level. 2) \textbf{ViCo-X}, a dataset of recorded dialogue scenarios by experienced actors with twenty-six different dialog acts, \ie communicative intentions, to emphasize the interaction part within multi-turn dialogues. Interlocutors would talk, listen and collaborate with real person to complete the conversation. The generation of conversational agents can be achieved from this.

Compared to speaker-centric datasets such as MEAD~\cite{wang2020mead}, VoxCeleb2~\cite{chung2018voxceleb2} and \etc, ViCo introduces the listener role to the conversation, focuses on the receipt and feedback of information and offers a vital complement to speaker-centeric tasks. It features real persons in actual conversations, thus enabling natural human interactions to reflect all three dimensions found in classical oral conversations. While ViCo-X records the behaviors of two interlocutors in multiple turns of a dialogue, paying more attention to the ``interactions''. The multi-turn dialogues are brought in as a video corpus, bringing the possibility of long-term mutual interaction modeling.

Based on ViCo and ViCo-X, we propose to formulate a conversational agent who aims to model one of the interlocutors in the conversation. It is designed as a digital twin of interlocutor as~\cref{fig:taskdescription} (d) shows. We further nominate two sub-tasks, expressive talking head generation and responsive listening head generation, to empower the agent with communication capabilities. Different from traditional talking head generation tasks that only transmit messages, expressive talking head generation introduces the listener role to speaker modeling, enabling speakers to notice the listener feedback concurrently, \ie, bi-directional communication. Responsive listening head generation focuses on listener modeling and targets to synthesize listeners that are responsive, \ie, dynamically respond to speaker verbal and non-verbal signals. 
And finally, to achieve head generation in multi-turn dialogues, a role transformer network is included to make interlocutors' role switches in a seamless and natural way. Hence, the conversational property in digital human generation tasks could be highlighted. It allows us to synthesize an interlocutor who can normally communicate with real people. This capability is undoubtedly critical to a wide range of applications, including virtual anchors, digital influencers, customer representatives, and digital avatars in Metaverse, wherever it involves interactive and conversational communication.

The main contributions are summarized as follows:
\begin{itemize}
    \item We construct two datasets ViCo and ViCo-X for listener modeling and conversational agent modeling.
    \item Based on the datasets, we introduce a new task, conversational head generation, including expressive talking head generation and responsive listening head generation, to mimic one interlocutor in a conversation.
    \item Furthermore, we introduce a baseline approach for synthesizing an interlocutor capable of engaging in speaker/listener interactions with real individuals. Our method is evaluated through comprehensive quantitative and qualitative analyses, including user studies, which substantiate its effectiveness and efficiency.
\end{itemize}

\begin{table*}[]
  \centering
  \small
  \caption{Comparison with other human conversation-related datasets.}
  \label{tab:dataset_comparison}
  \resizebox{\textwidth}{!}{
  \begin{tabular}{@{}lccccccccc@{}}
    \hline\noalign{\smallskip}
    Dataset & Year & Public & Interlocutor & Multi-turn & Style & Environment & Head motion & Body motion & Other Anno \\
    \hline
    GRID      & 2006 & \cmark & Speaker & \xmark & Lab  & Realistic & \xmark & \xmark & - \\
    LRW       & 2016 & \cmark & Speaker & \xmark & Wild & Realistic & \xmark & \xmark & - \\
    ObamaSet  & 2017 & \cmark & Speaker & \xmark & Wild & Realistic & \cmark & \xmark & - \\
    VoxCeleb  & 2017 & \cmark & Speaker & \xmark & Wild & Realistic & \cmark & \xmark & - \\
    VoxCeleb2 & 2018 & \cmark & Speaker & \xmark & Wild & Realistic & \cmark & \xmark & - \\
    LRS2-BBC  & 2018 & \cmark & Speaker & \xmark & Wild & Realistic & \cmark & \xmark & - \\
    LRS2-TED  & 2018 & \cmark & Speaker & \xmark & Wild & Realistic & \cmark & \xmark & - \\
    Faceforensics++ & 2019 & \cmark & Speaker & \xmark & Wild & Realistic & \cmark & \xmark & - \\
    MEAD      & 2020 & \cmark & Speaker & \xmark & Wild & Realistic & \cmark & \xmark & emotion \\
    \hdashline[.4pt/1pt]
    Speech2Gesture & 2019 & \cmark & Presenter & \xmark & Wild & Realistic & \cmark & \cmark & - \\
    Ted Gesture    & 2019 & \cmark & Presenter & \xmark & Wild & Realistic & \cmark & \cmark & - \\
    \hdashline[.4pt/1pt]
    Gillies~\etal~\cite{gillies2008responsive} & 2008 & \xmark & Speaker, Listener & \xmark & Lab & Simulated & \cmark & \cmark & - \\
    SEMAINE~\cite{mckeown2011semaine} & 2011 & \cmark & Speaker, Listener & \xmark & Lab & Simulated & \cmark & \xmark &   custom dimension \\
    Heylen~\etal~\cite{heylen2011generating} & 2011 & \xmark & Speaker, Listener & \xmark & Lab & Simulated & - & - & - \\
    ALICO~\cite{buschmeier2014alico} & 2014 & \xmark & Speaker, Listener & \xmark & Lab & Realistic & \cmark & \xmark &  feedback signal \\
    \hdashline[.4pt/1pt]
    ViCo & 2022 & \cmark & Speaker, Listener & \xmark & Wild & Realistic & \cmark & \xmark & attitude \\
    ViCo-X & 2023 & \cmark & Conversational Agent & \cmark & Lab & Realistic & \cmark & \cmark & dialogue act \\
    \hline
  \end{tabular}
  }
\end{table*}

\section{Related Works} \label{sec:related}
\textbf{Speaker-centered video synthesis}\quad Given time-varying signals and a reference still image of the speaker, the talking head synthesis task aims to generate a vivid clip for the speaker with the time-varying signals matched. Based on the different types of time-varying signals, we can group these tasks into two groups: 1) audio-driven talking head synthesis~\cite{chung2017you,prajwal2020lip,zhang2021facial,yu2021multimodal,eskimez2021speech, doukas2023free}, 2) video-driven talking head synthesis~\cite{bansal2018recycle}. The goal of the former one is to generate a video of the speaker that matches the audio. And the latter one is to generate videos of speakers with expressions similar to those in the video. They only matter to transmit messages to listeners while ignoring listener feedback. Our expressive talking head generation is an improved version: we do not just talk, we communicate.

\textbf{Listening behaviors modeling}\quad Many applications and research papers have focused on speaking, while ``listener modeling'' is seldom explored. Gillies \etal~\cite{gillies2008responsive} first propose the data-driven method that can generate an animated character that can respond to speaker's voice. This lacks the supervision of speaker visual signals, which is incomplete for responsive listener modeling. And this method can not be applied to realistic head synthesis. Heylen \etal~\cite{heylen2011generating} further studied the relationship between listener and speaker audio/visual signals from a cognitive technologies view. SEMAINE~\cite{mckeown2011semaine} records the conversation between a human and a limited artificial listener. MAHNOB Laughter database~\cite{petridis2013mahnob} focuses on studying laughter's behaviors when watching funny video clips. Apart from these related works, ALICO~\cite{buschmeier2014alico} corpus about active listener analysis is the most relevant dataset to our proposed task. However, it has not been made public and also not constructed from the real scene conversations. Moreover, the main objective of ALICO is for psychology analysis, the data mode of that dataset is vastly different from the audio-video corpora in computer vision area. In the past few years, social AI intelligence~\cite{joo2019towards,oertel2021towards} has been introduced to model the nonverbal social signals in triadic or multi-party interactions. Joo~\etal~\cite{joo2019towards} is concerned with the overall posture and head movement of a person, and Oertel~\etal~\cite{oertel2021towards} aims to mine listening motion rules for robotics controlling. Both related works only deal with the speaking status and ignore the speaker's content. What's more, they rarely care about two-person interactions nor pay attention to model the face in detail, which is also different from our task. A detailed comparison to existing related datasets is shown in~\cref{tab:dataset_comparison}.

As far as we know, this is the first time to introduce the learning-based listening head generation task in computer vision area. In this work, we propose a formulation of responsive listening head generation and construct a public ViCo dataset for this task. Meanwhile, a baseline method is proposed for listening head synthesis by perceiving both speaker's audio/visual signals and preset attitude.

\textbf{Speaker-Listener Coupling in Communication}\quad Classical two-party face-to-face communication usually includes two interlocutors' collaboration. They would act as speakers or listeners in turn and finally agree with some points, then finalize the conversation. We can notice the "coupling" in this procedure: each interlocutor plays both the speaker and listener roles. This phenomenon is traceable to the 
neuroscience~\cite{stephens2010speaker} level and is deserved to be studied for more realistic and emotional digital human modeling. 

\section{Task Overview} \label{sec:overview}
We present Interactive Conversational Head Generation, a novel task involving the synthesis of one interlocutor's head during the conversation. In particular, our aim is to comprehend the dialog act and the behaviors of other interlocutors, such as audio cues, head motions, facial expressions, eye blinks, and more.

Assume there are two interlocutors $\cP$ and $\cQ$, given the input $N$-turn video sequences of $\cQ$: $\cV^{\cQ}_{N}=\{\biV^\cQ_1, \biV^\cQ_2, \dots, \biV^\cQ_N\}$, these two interlocutors' audio signal sequences $\cA=\{\biA_1, \biA_2, \dots, \biA_N\}$, the speaker's dialog acts or listener's attitudes $\biE=\{e_1, e_2, \dots, e_N\}$, and the role indicator vector $\biR=\{r_1, r_2, \dots, r_N\}$ ($r_i=1$ for speaker and $r_i=0$ for listener) of $\cP$, the conversational head generation task aims to generate the visual representations of $\cP$ in this $N$-turn conversation that can both listen and talk in one paradigm:
\begin{equation}
    \cV^{\cP}_N = \bfG(\cA, \biR, \biE, \cV^\cQ_N, v^{\cP}),
\end{equation}
where $v^{\cP}$ represents the initial (identity) image of the interlocutor $\cP$. Note that the $\cV^{\cdot}_{i}$ represent a video clip of this interlocutor and can be further subdivided as $\cV^{\cdot}_{i}=\{\biV^{\cdot}_{i;1}, \biV^{\cdot}_{i;2}, \dots, \biV^{\cdot}_{i;|\biA_i|}\}$, where $\biV^{\cdot}_{i;j}$ denotes the image.

\textbf{Speaker dialog act definition} Dialog act theory provides an empirically-grounded framework for computational modeling of communication, specifically focusing on linguistic and nonverbal behaviors within a dialogue. Informally speaking, dialog acts are such actions as \textit{providing information}, \textit{apologizing for a misunderstanding}, and \textit{giving feedback}, \etc. This theory remains independent of any specific dialog system and effectively captures the actions performed by the interlocutors. Roughly, it can be divided into 9 core orthogonal dimensions (\eg, task, turn management, time management, social obligations management, own/partner communication management and \etc) and hosts total of 57 functions. These dialog acts are theoretically justified and domain-independent. They can also be recognized by human analysts from the ``semantic'' view.

\textbf{Listener attitude definition}\quad During conversation, after perceiving the signals from the speaker, the listener usually reacts with an active, responsive \emph{attitude}, including epistemic attitudes (\eg, agree, disagree) and affective attitudes (\eg, like, dislike). In general, attitude potentially guides the listener's behavior and consequently affects the conversation. Also, different attitude results in different facial expressions and behaviors of the listener~\cite{heylen2007searching}, e.g. a laugh appears as the most appropriate signal for \textit{like}, a combination of smile and raise eyebrows could be a possibility for \textit{interested}, \textit{disagree} can be meant by a head shake, \textit{dislike} is represented by a frown and tension of the lips, \etc.

\textbf{Feature extraction}\quad
In this work, we extract the energy feature, temporal domain feature, and frequency domain feature of the input audio; and model the facial expression and head poses using 3DMM~\cite{blanz1999morphable} coefficients.

For the audio, we extract the Mel-frequency cepstral coefficients (MFCC) feature with the corresponding MFCC Delta and Delta-Delta features. Furthermore, we incorporate the energy, loudness, and zero-crossing rate (ZCR) are also embedded into audio features denoted as $\biS_{i}$ for each audio clip $\biA_{i}$. The audio feature extracted from $\cA$ can be denoted as $\cS=\{\biS_{1}, \biS_{2}, \dots,\biS_{t}\}$.

There has been extensive research~\cite{tewari2018high, tran2019learning, booth20183d} in the area of 3D face reconstruction. Here, we leverage the state-of-the-art deep learning-based face reconstruction model~\cite{deng2019accurate} to get the 3DMM~\cite{blanz1999morphable} coefficients and then drive the head movements and expression changes. Specially, for any facial image, we can get the reconstruction coefficients $\{\alpha, \beta, \delta, p, \gamma\}$ which denote the identity, expression, texture, pose and lighting, respectively. Further, we distinguished the 3D reconstruction coefficients into two parts: $\FixedMM=(\alpha, \delta, \gamma)$ to represent relatively fixed, identity-dependent features, and $\DynamMM=(\beta, p)$ to represent relatively dynamic, identity-independent features. After the face parameterization, we can adjust the motion and expression changes through $m$, and modify the identity through $\cI$.

\textbf{Task definition}\quad Given that the identity-dependent features ($\cI$) are weakly correlated with the interlocutor motion patterns, we use only the head motion and facial expression feature $m$ for conversational head generation model training and then adapt the identity-dependent features $\cI$ of different interlocutor identities for visualization and evaluation. Thus our interactive conversational head synthesis task can be formulated:
\begin{equation} \label{eq:conversation}
\begin{split}
    \cM^\cP_N &= \bfG_m(\cA, \biR, \biE, \cM^\cQ_N, m^\cP), \\
    \cV^\cP_N &= \bfG_v(\cM^\cP_N, \cI^\cP, v^\cP), \\
\end{split}
\end{equation}
where $\cM^\cdot_i$ is the dynamic feature predicted for this interlocutor of $i$-th round. Here, $\bfG_m$ would infer 3DMM coefficients of $\cP$ and $\bfG_v$ render the coefficients into video. To get a better modeling for $\bfG_m$, we split it into expressive talking head modeling and responsive listening head modeling with respect to $\biR$ and then combine them to formulate the whole conversational head. 

Start with a much simpler subproblem, considering a single turn of this conversation. Our aim is either to generate a listener based on the speaker's audio $\biA$, behaviors $\cM^\cS$ and current dialog act $e$:
\begin{equation} \label{eq:listening}
    \cM^\cL = \bfG_m^\cL(\biA, e, \cM^\cS, m^\cL),
\end{equation}
where $\cS$ and $\cL$ denote the speaker and listener respectively, or to generate a speaker based on the audio $A$, dialog act or attitude $e$ and listener behaviors $\cM^\cL$:
\begin{equation} \label{eq:talking}
    \cM^\cS = \bfG_m^\cS(\biA, e, \cM^\cL, m^\cS),
\end{equation}
For the former task, we propose a framework for responsive listening head generation, which aims to fill the blank of listener modeling. And for the latter task, we introduce listener behaviors $\cM^\cL$ to classical talking head generation, named expressive talking head generation, to make speaker better communicate with listener.

And finally, towards generation in $N$-round and ensure the interlocutors' role switches are smooth, we continue to solve~\cref{eq:conversation} in conversational wide to modeling agent $\cP$.

The 3D face rendering technology $\Render$ has been well studied in many recent works~\cite{kim2018deep,zhang2021facial}. Moreover, the face rendering models are usually identity-specific, so one may need to train separately for each identity for better performance. To highlight the properties of the conversational head synthesis task, and decouple the critical factor, our proposed model primarily focuses on the motion-related and identity-independent 3D facial coefficients prediction task $\mathbf{G}_m$, and use the pretrained rendering  model~\cite{ren2021pirenderer} for simplified visualization. And some video post-processing methods, such as video frame interpolation~\cite{bao2019memc}, denoising~\cite{chen2021multiframe}, super resolution~\cite{dong2015image,yi2020progressive}, in-painting~\cite{szeto2019temporally}, can be used for better visual effects.

\section{Dataset Construction}
\label{sec:dataset}
\subsection{ViCo Dataset}
\begin{figure*}[t]     
    \centering
    \includegraphics[width=\linewidth,page=1]{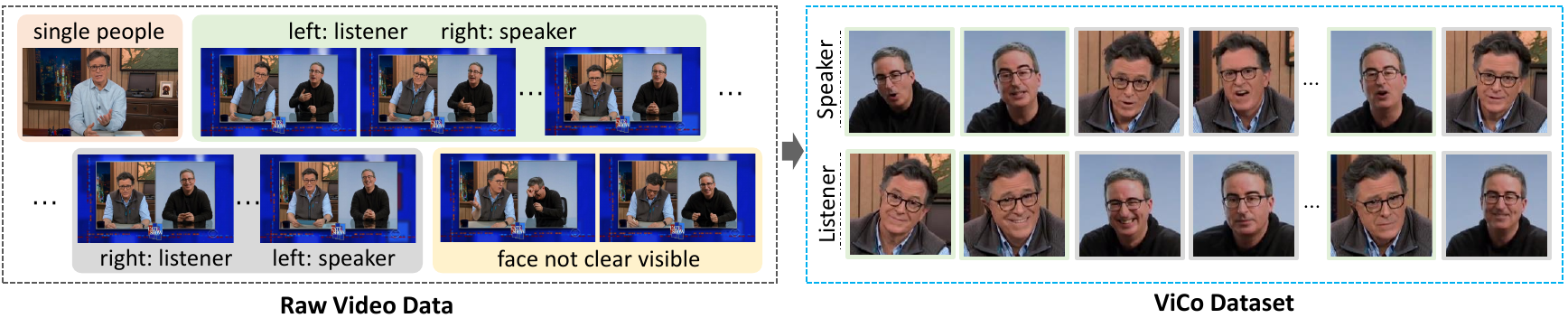}
    \caption{In ViCo, valid clips are selected in accordance with the standards that 1) both the speaker and listener behaviors are clearly visible, and 2) listeners are responsively engaged to the conversation. The facial regions of listener-speaker pairs are further cropped for constructing our ViCo dataset (right).}
    \label{fig:dataset_construct}
\end{figure*}

To highlight the ``listener'' dimension, we first construct a speaker-listener dataset ViCo mainly for responsive listening-head generation by capturing \emph{conversational} video clips from YouTube containing two people's frontal faces. To ensure the validity of a video clip, it must fulfill the following conditions::
\begin{itemize}[itemsep=2pt,topsep=2pt,parsep=2pt]
    \item The screen should display only two individuals, with one person actively speaking and the other attentively listening.
    \item Both individuals' frontal faces should be prominently visible, ensuring clear visibility.
    \item The facial expressions of both individuals should appear natural and stable throughout the clip.
    \item The listener should actively engage with the speaker, responding dynamically and in real time to foster an interactive exchange.
\end{itemize}

The annotators were tasked with meticulously documenting the precise start and end times of each ``valid'' clip. Additionally, they were instructed to label the speaker's position (left or right of the screen) and discern the listener's attitude exhibited in the video. In ViCo, we group the attitudes into three categories: positive, negative and neutral. Positive attitude consists of \textit{agree}, \textit{like}, \textit{interested}. Conversely, negative attitude consists of \textit{disagree}, \textit{dislike}, \textit{disbelieve}, \textit{not interested}. Cross-validation was applied among at least three annotators for each candidate clip for quality control. For each valid clip, we use the MTCNN to detect the face regions in each frame, and then crop and resize the detected face regions to 384$\times$384 resolution image sequence for model training and evaluation, as shown in~\cref{fig:dataset_construct}.

\setlength{\tabcolsep}{4pt}
\setlength\intextsep{8mm}
\begin{table}[t]
  \centering
  \small
  \cprotect\caption{Statistics of ViCo dataset. \#ID indicates the number of identities. The same person/identity can play different roles with multiple attitudes.}
   \label{tab:dataset_statistics}
  \resizebox{\linewidth}{!}{
  \begin{tabular}{@{}lcccccc@{}}
    \hline\noalign{\smallskip}
    Attitude & \#Videos & \#Speaker & \#Listener & \#ID & \#Clips & Duration \\
    \noalign{\smallskip}
\hline
\noalign{\smallskip}
    Positive & 42 & 53 & 62 & 81 & 226 & \SI{49}{\minute}~\SI{18}{\second} \\
    Neutral  & 35 & 38 & 48 & 63 & 134 & \SI{27}{\minute}~~\SI{7}{\second} \\
    Negative & 11 & 11 &  9 & 18 &  123 & \SI{18}{\minute}~\SI{57}{\second} \\
    \hline
    Total    & 50 & 67 & 76 & 92 & 483 & \SI{95}{\minute}~\SI{22}{\second} \\
    \hline
  \end{tabular}
  }
\end{table}
\setlength\intextsep   {8mm}

\Cref{tab:dataset_statistics} shows the statistical information of our annotated responsive listening head generation dataset ViCo. The proposed dataset contains rich samples of 483 video clips. We normalize all videos to 30 FPS. Moreover, our dataset is composed of high-quality videos (1920$\times$1080) and audios (\SI{44.1}{\kHz}/\SI{16}{bit}) and contains diverse scenarios, including news interviews, entertainment interviews, TED discussions, variety shows and \etc, which provides rich semantic information and various listener patterns. The video clips' length can range from \SI{1} to \SI{71} seconds.

\subsection{ViCo-X Dataset}
\begin{figure}
    \centering
    \includegraphics[width=\linewidth]{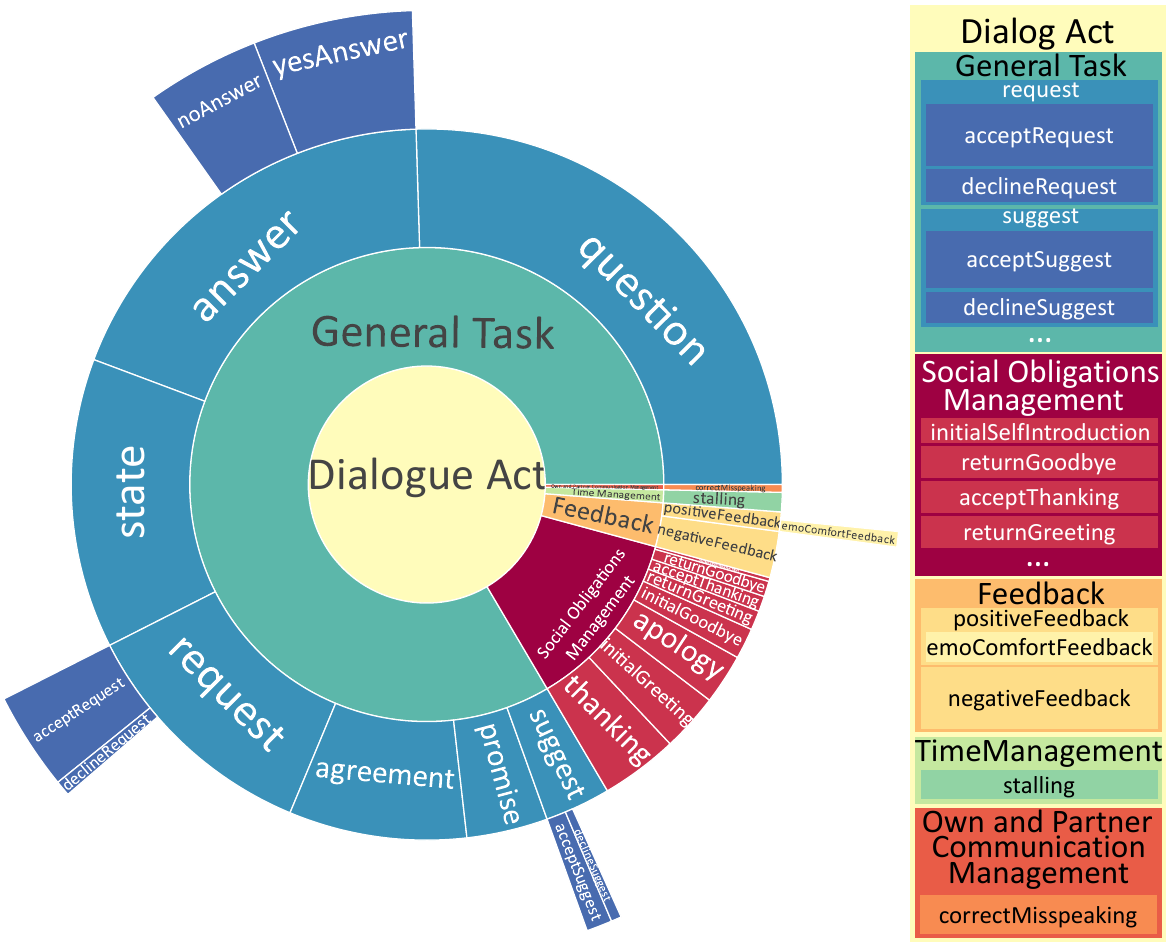}
    \caption{Dialog act distribution in ViCo-X dataset.}
    \label{fig:da_distribution}
\end{figure}
We have collected the in-the-wild dataset ViCo for talking head generation and listening head generation. However, limited by the video quality, conversation content, and scenario restrictions, the functional multi-turn conversational data is still unavailable. Thus we use higher quality devices to record conversational ``interactions'' in specific scenarios to build the multi-turn conversation dataset, ViCo-X. 

To the best of our knowledge, this is the first multi-turn multimodal interaction dataset (ViCo-X) to model human behaviors in conversation. The dataset consists of data in three modalities: audio, video, and text, and a fine-grained dialog act annotation is attached to each sentence in the conversation. We establish this multi-modality dataset to encourage more studies in ``interactive face-to-face conversation understanding and modeling''.

In this dataset, we focus on the multi-turn conversations in a functional and application-oriented scenario, thus using the Jing Dong Dialogue Corpus (JDDC)~\cite{chen2019jddc}, the large-scale real scenario Chinese E-commerce conversation corpus, as the transcriptions. Based on this corpus, we first filter out the sensitive and inappropriate dialogues. After that, we select representative and diverse dialogues, and further label the dialog acts (DA) following the ISO standard~\cite{bunt2012iso}, and its distribution in our dataset is shown in~\cref{fig:da_distribution}. To ensure rigorous quality control in DA labeling, cross-validation was implemented using a minimum of three annotators with the standard of Fleiss Kappa exceeding 0.9.

\begin{figure}
    \centering
    \includegraphics[width=0.9\linewidth]{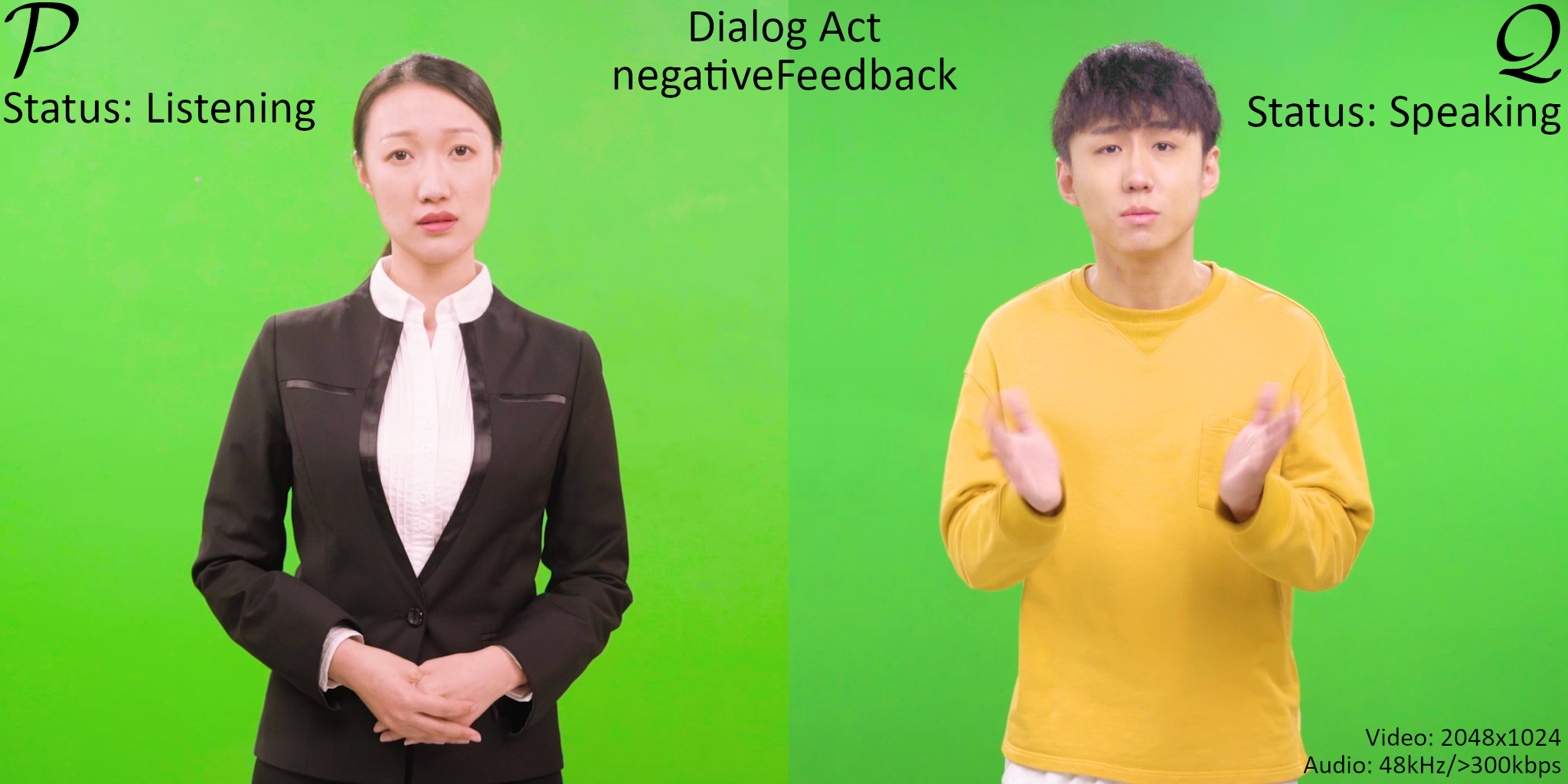}
    \caption{A screenshot of conversational interaction between two interlocutors in the recording.}
    \label{fig:dataset_example}
\end{figure}

Then, given the transcriptions, during the recording process, the actors are spaced two meters apart, with their bodies facing the camera and their backs against the green screen. Actors are asked to communicate messages and intentions through words, expressions, head movements, hand postures, or body gestures under the dialog act constraints. Speakers are required to raise their emotions and speak their transcripts in a natural style, and listeners should be responsive and active to the speaker. There is a guidance team to guarantee the recording quality and the professionalism of actors. \Cref{fig:dataset_example} shows an image when recording. Our videos are in $2048\times1024$ and 30 fps format, and audios are \SI{48}{\kHz} and over \SI{300}{kbps}. Thus the details can be maintained well.

With the recording finished annotators are asked to indicate the start and end time of each sentence (accurate to $\sfrac{1}{30}$ second) and label the position of the speaker (left or right of the screen). Finally, these components make up the entire ViCo-X dataset totaling 44k frames.

% Compared to ViCo dataset, alongside the improved real-time, our recorded ViCo-X is of a higher quality and is established on the real E-commerce conversation corpus. The twenty-five different scenarios can cover most of the conversation between customer service and customers and provide rich semantic information and various conversation patterns. It can ensure help studies of human-human interaction in specific scenarios and suggest implications for the mutual interaction between the virtual human and the real human. 
In contrast to the ViCo dataset, our newly recorded ViCo-X dataset not only offers enhanced real-time performance but also boasts higher-quality recordings. ViCo-X is built upon a real E-commerce conversation corpus, encompassing twenty-five diverse scenarios that encompass the majority of interactions between customer service representatives and customers, which provides a wealth of semantic information and encompasses various conversation patterns. It enables comprehensive studies of human-human interactions within specific scenarios and offers valuable insights into the dynamics of mutual interaction between virtual humans and real humans.

In contrast to the existing talking head video datasets~\cite{chung2018voxceleb2, wang2020mead} and listening head video dataset ViCo that aims to model a sub-part of the conversation, ViCo-X takes a different viewpoint by focusing on modeling the real conversations. It emphasizes the dynamic interaction between interlocutors who can alternate roles as both speakers and listeners, and prioritizes mutual engagement rather than unidirectional information perception or conveyance.

% Dissimilar from the existing talking head video datasets~\cite{chung2018voxceleb2,wang2020mead} and listening head video dataset, ViCo, which aims to model a sub-part of the conversation, ViCo-X focuses on the conversation modeling in real conversation, that is, the interlocutor can act as both the speaker and listener. It matters about mutual interaction rather than a unidirectional information perception or conveyance.

\section{Interactive Conversational Head Generation} \label{sec:conversational_generation}
This section demonstrates how to achieve interactive conversational head generation in multi-turn communicative scenarios as outlined in~\cref{sec:overview}.

\subsection{Responsive Listening Head Generation} \label{sec:listening_head}
According to psychological knowledge, an active listener tends to respond based on the speaker's audio~\cite{kendon1970movement} and visual signals~\cite{gillies2008responsive,maatman2005natural}. And at a given moment, the listener receives information from the speaker of that moment as well as information from history and adopts a certain attitude to present actions in response to the speaker. Thus, the goal of our model is to estimate the conditioned probability $P(\cM^\cL \vert \biA, e, \cM^\cS, m^\cL)$, where the $\cM^\cS$ and $\biA$ are time-varying signals that the listener should respond to, and the reference listener feature $m^\cL$ and attitude $e$ constrain the pattern of the entire generated sequence.

Inspired by the sequence-to-sequence model, a multi-layer sequential decoder module $\mathbf{G}_m$ is applied for modeling the time-sequential signals of conversation. Unlike talking-head generation~\cite{chung2017you,wang2020mead,zhu2021deep}, which accepts an entire input of audio and then processes it using a bidirectional LSTM or attention layer; in our scenario, the model $\Generator$ receives the streaming input of the speaker where future information is not available.

For the speaker feature encoder, at each time step $t$, we first extract the audio feature $\AudioFeat{t}$ and the speaker's head and facial expression representation $\SpeakerFeat{t}$, then apply non-linear feature transformations following a multi-modal feature fusion function $f_{am}$ to get the encoded feature of speaker. The representation of reference listener $\ListenerFeat{1}$ and attitude $e$ can be embedded as the initial state $h_1$ for the sequential motion decoder. At each time step $t$, taking the speaker's fused feature $f_{am}(\AudioFeat{t}, \SpeakerFeat{t})$ as input, $\bfG^\cL_m$ in Eq.~\ref{eq:listening} is functioned as updating current state $h_{t+1}$ and generating the listener motion $m_{t+1}^l$, which contains two feature vectors, \textit{i.e.} $\beta_{t+1}^l$ for the expression and $p_{t+1}^l$ for the head rotation and translation. Our responsive listening head generator supports an arbitrary length of speaker input. The procedure can be formulated as:
\begin{equation}
    \beta_{t+1}^l, p_{t+1}^l = \mathbf{G}^{\cL}_m(h_t, f_{am}(s_t, m_t^s)).
\end{equation}
This way, the responsive listener can be synthesized by referring to the speaker's verbal and non-verbal signals.

\subsection{Expressive Talking Head Generation} \label{sec:talking_head}
Talking head generation aims to synthesize a clear and vivid talking-head video that the identity matches the given reference image(s) and the motion (notably for lips) corresponds to the driven source (\eg, a piece of audio speech). It has been well studied in recent years, a classical 3DMM coefficients-based solution can be formulated as:
\begin{equation} \label{eq:talking_detail}
    \beta^{s^*}_{\{1,\cdots,t\}}, p^{s^*}_{\{1,\cdots,t\}} = \bfG^{\cS^*}_m(\bfA, e, m^\cS),
\end{equation}
where $e$ could be some factors those affect speaker behaviours, \eg, emotion, dialogue act, \etc., while such efforts are neglected the listener. From the perspective of psychology and social behavior, the speakers should also react to the listener's non-verbal response when talking. Here we propose the listener-aware speaker modeling, which targets making the speaker more lively with the listener supervision by introducing the concept of ``interactive'' to the speaker. 

As~\cref{fig:taskdescription}(d) left subfigure shown, different from~\cref{fig:taskdescription}(c) which the driven source is audio signals only, we force the talking head generation model to receive the streaming input of listener frames and expect the synthesized video more suitable for face-to-face conversation scenario. 

In view of the given audio are batched and can be modeled with bi-directional models~\cite{zhang2021facial} while the listener inputs are streaming, we divide the talking head modeling into two parts: audio-supervised speaker modeling and listener-supervised speaker modeling, and fuse these two for final results. The former task has been well studied in previous literature, and the latter task is similar to~\cref{sec:listening_head}: At each time step $t$, we would extract the listener's representation $m_t^l$ and encode the representation of reference speaker $m_t^s$ and $s$ as the initial state $h'_1$ for the listener-aware speaker decoder. Then we can predict the expected speaker head motion and expression which reflect to the listener:
\begin{equation}
    \beta_{t+1}^{s'}, p_{t+1}^{s'} = \bfG_m^{\cS'}(h'_t, m_t^l)
\end{equation}
Then a weighted fusion is performed for blending the output of audio-supervised speaker model and listener-supervised speaker model:
\begin{equation}
\begin{aligned}
    \beta_{t+1}^s &= \alpha_\beta \beta_{t+1}^{s'} + (1 - \alpha_\beta) \beta_{t+1}^{s^*}, \\
    p_{t+1}^s &= \alpha_p p_{t+1}^{s'} + (1 - \alpha_p) p_{t+1}^{s^*},
\end{aligned}
\end{equation}
where $\alpha_\beta$ and $\alpha_p$ are trainable parameters that can be optimized during training procedure.

\subsection{Interactive Conversational Head Generation} \label{sec:conversational_agent}
In the real scenario, a conversation is usually multi-turn and each interlocutor can act as both the speaker and listener. Existing works mostly focus on modeling a single role and neglect the \textit{constant transformation} for each interlocutor between speaker and listener. To tackle this problem and towards face-to-face conversation modeling, we propose a novel task: conversational head generation, which aims to synthesize a conversational head which can interact with human avatar in both talking and listening manner and the state can be well switches during role alteration as illustrated in~\cref{fig:taskdescription}(d). 

Instead of using if/else toggles directly, which would make the transformation rigid and spiky, we introduce a more smooth transformation trick to switch the agent role. To alternate the agent role from $r_{t-1}$ to $r_{t}$ ($r_{t-1}\neq r_{t}$), the hidden state $h_{t}$ is inferred from the previous state $h_{t-1}$:
\begin{equation}
    h_t = \bfT(h_{t-1}, r_{t-1}, r_{t}),
\end{equation}
where $T$ is role switcher network to bridge $r_{t}$ with $r_{t-1}$. In such manner, we can extend the single-turn speaker/listener modeling to multi-turn agent modeling, formulating a complete conversational head.

\subsection{Optimization Objectives}
For optimization, regardless of the role in the communication, we use a consistent optimization strategy. With the ground truth patterns denoted as  $\RoleFeatsHatT=[\RoleFeatHat{2}, \RoleFeatHat{3}, \cdots, \RoleFeatHat{T}]$, we drop the last prediction $\RoleFeat{T+1}$ due to the lack of supervision and use $L_2$ distance to optimize training procedure:
\begin{equation}
    \cL_{gen} = \sum_{t=2}^{T} \Vert \beta_t - \hat{\beta}_t \Vert_2 + \Vert p_t - \hat{p}_t \Vert_2.
\end{equation}

Moreover, a motion constraint loss $\cL_{mot}$ is applied to guarantee the inter-frame continuity across $\RoleFeatsHatT$ is similar to the predicted $\RoleFeatsT$:
\begin{equation}
    \cL_{mot} = \sum_{t=2}^{T} w_1\Vert \mu(\beta_t) - \mu(\hat{\beta}_t) \Vert_2 + w_2\Vert \mu(p_t) - \mu(\hat{p}_t) \Vert_2,
\end{equation}
where $\mu(\cdot)$ measures the inter-frame changes of current frame and its adjacent previous frame, i.e., $\mu(\beta_t)=\beta_t-\beta_{t-1}$, $w_1$ and $w_2$ is a weight to balance the motion constraint loss and generation loss. The final loss function of this communication role can be formulated as:
\begin{equation}
    \cL_{total} = \cL_{gen} + \cL_{mot}.
\end{equation}

For the responsive listening head generation (\cref{sec:listening_head}) and expressive talking head generation (\cref{sec:talking_head}), we optimize the $\cL_{total}$ for the listener and speaker patterns respectively. And for the conversational head finetune (\cref{sec:conversational_agent}), we optimize these two patterns simultaneously.
\section{Experiments}
\subsection{Implementation Details}
The ViCo dataset is divided into two parts: \RomanNumeral{1}) training set for learning listener / speaker patterns, \RomanNumeral{2}) test set for validating the performance, generalizability, and transferability of our model. All identities do not overlap between train set and test set. And in ViCo-X dataset, we perform a five-fold cross validation and results are conducted on the whole dataset.

We extract 45-dimensional acoustic features for audios, including 14-dim MFCC, 28-dim MFCC-Delta, energy, ZCR and loudness. There are multiple choices to implement $\mathbf{G}_m$, such as standard sequential models like LSTM, GRU, or a Transformer decoder with sliding window. Here we adopt LSTM for our baseline since it has been widely used in many similar applications such as motion generation~\cite{richard2021meshtalk}, and achieve stable state-of-the-art performance when training on small corpus~\cite{melis2019mogrifier}. The role switcher network $\bfT$ is implemented just by two linear mappings (listener$\leftrightarrow$speaker) on hidden states. Our models are trained with AdamW optimizer with a learning rate of \SI{2e-3}{} (decayed exponentially by 0.5 every 30 epochs), $\beta_1=0.9$ and $\beta_2=0.999$, for 300 epochs. For all experiments, we keep the same hyper-parameter for fair and set $w_1$ to \SI{1e-3}{} and $w_2$ to 1.

\begin{table*}\centering
\caption{Results of Responsive Listening Head Generation on ViCo and ViCo-X datasets.}
\label{tab:listening_results}
\begin{tabular}{@{}llcccccccccc@{}}\toprule 
\multirow{2}{*}{Dataset} & \multirow{2}{*}{Method} & \multicolumn{3}{c}{Generator} & \phantom{a} & \multicolumn{5}{c}{Render} \\
\cmidrule{3-5} \cmidrule{7-11}
&& ExpFD $\downarrow$ & AngleFD $\downarrow$ & TransFD $\downarrow$ && SSIM $\uparrow$ & CPBD $\uparrow$ & PSNR $\uparrow$ & FID $\downarrow$ & CSIM $\uparrow$ \\ 
\midrule
\multirow{3}{*}{ViCo} & Random & 16.435 & 9.401 & 7.600 && 0.592 & 0.104 & 17.497 & 47.751 & 0.060 \\
& Mirror & 18.882 & 10.155 & 7.624 && 0.553 & 0.086 & 16.224 & 61.280 & 0.063 \\
& Responsive & \textbf{15.149} & \textbf{7.963} & \textbf{6.721} && 0.584 & 0.097 & 17.492 & 52.947 & 0.058 \\
\midrule
\multirow{3}{*}{ViCo-X} & Random & 13.739 & 5.913 & 6.316 && 0.654 & 0.154 & 18.615 & 37.277 & 0.176 \\
& Mirror & 17.635 & 7.744 & 8.844 && 0.630 & 0.172 & 17.523 & 48.286 & 0.184 \\
& \textcolor{TRed}{Responsive} & \textbf{10.656} & \textbf{4.350} & \textbf{5.129} && 0.685 & 0.183 & 20.032 & 36.707 & 0.194 \\
\bottomrule
\end{tabular}
\end{table*}

\subsection{Evaluation Metrics}
Since we use a detached renderer rather than an end-to-end pipeline, we can divide the assessment into two sides: the performance of listener generator $\bfG_m$ and the visual effects of renderer $\bfG_v$. For the former one (the \textbf{Generator} Part), we use $L_1$ distance between the generated features and the ground-truth features (FD) to ensure the predicted fine-grained head and expression coefficients similar to the ground-truth. And for the latter one (the \textbf{Renderer} Part), we select the Fr\'echet Inception Distance (FID)~\cite{heusel2017gans}, Structural SIMilary (SSIM), Peak Signal-to-Noise Ratio (PSNR) and Cumulative Probability of Blur Detection (CPBD) to evaluate the visual effects of renderer. Besides, the cosine similarity from ArcFace is introduced for identity-preserving measurement. And for talking head generation, we apply the Landmark Distance (LMD) for the fine lip movements matching, and the lip-sync metric by SyncNet~\cite{chung2016out} for the synchronization of lip motion with the input audio. Since we did not contribute to the renderer part and instead used a pre-trained renderer, \textit{we only assess the model performance with the Generator Part and leave the Renderer Part to provide a basic criterion for evaluation of future realistic face rendering research work}.

In order to properly evaluate the performance of the conversational head generation system, it is necessary to conduct user studies. To ensure fair results, 20 volunteers will be recruited to rate the synthesized results.

\subsection{Quantitative Results}
\textbf{Responsive Listening Head Generation}\quad In~\Cref{tab:listening_results}, we report the results across different listening head generation methods, including 1) ``Random'': generate frames from reference image but injecting small perturbations in a normal distribution to mimic random head motion. 2) ``Mirror'': copying the speaker's motion patterns $\cM^S$. 3) ``Responsive'': the proposed responsive listening head generation method (\cref{sec:listening_head}). 

The results show that in both datasets, ViCo and ViCo-X, the ``Responsive'' listeners significantly outperformed the traditional non-parametric listeners (``Random'' and ``Mirror''). This significant difference in performance can be attributed to the introduction of speakers' audio and visual signals. This is also in line with our psychological perception that the listener doesn't just do some mechanical or meaningless actions, it coordinates with the speaker's behaviors in real-time and dynamically.

\textbf{Expressive Talking Head Generation}\quad \Cref{tab:talking_results} shows how the listener's visual signals affect speaker behaviors. In ViCo and ViCo-X, the audio + listener speaker shows a similar trend of improvements compared to audio-only speakers. With the integration of listener visual signals, the speakers can exhibit more realistic expressions. These findings support that communication and presentation are distinct, and monologues are not leading to communication.

\begin{table*}\centering
\caption{Results of Expressive Talking Head Generation on ViCo and ViCo-X dataset.}
\label{tab:talking_results}
\begin{tabular}{@{}llcccccccccccccc@{}}\toprule 
\multirow{2}{*}{Dataset} & \multirow{2}{*}{Method} & \multicolumn{6}{c}{Generator} & \phantom{a} & \multicolumn{5}{c}{Render} \\
\cmidrule{3-8} \cmidrule{10-14}
&& ExpFD $\downarrow$ & AngleFD $\downarrow$ & TransFD $\downarrow$ & LMD $\downarrow$ & AVOffset $\downarrow$ & AVConf $\uparrow$ && SSIM $\uparrow$ & CPBD $\uparrow$ & PSNR $\uparrow$ & FID $\downarrow$ & CSIM $\uparrow$ \\ 
\midrule
\multirow{2}{*}{ViCo} & Audio Only & 17.486 & \textbf{7.642} & 7.450 & \textbf{13.505} & 0.376 & 1.480 &&  0.603 & 0.080 & 17.977 & 45.581 & 0.086 \\
& Audio + Listener & \textbf{16.604} & 8.302 & \textbf{6.641} & 14.366 & \textbf{0.094} & \textbf{1.948} && 0.586 & 0.073 & 17.426 & 46.181 & 0.085 \\
\midrule
\multirow{2}{*}{ViCo-X} & Audio Only & 12.731 & \textbf{5.357} & 5.271 & \textbf{7.744} & 0.636 & 2.524 && 0.673 & 0.183 & 19.362 & 35.111 & 0.211 \\
& \textcolor{TBlue}{Audio + Listener} & \textbf{12.690} & 5.557 & \textbf{5.192} & 8.100 & \textbf{0.083} & \textbf{2.755} && 0.668 & 0.180 & 19.123 & 35.850 & 0.214 \\
\bottomrule
\end{tabular}
\end{table*}

\begin{table*}\centering
\caption{Results of Conversational Head Generation on ViCo-X dataset.}
\label{tab:agent_resultsx}
\begin{tabular}{@{}lccccccccccccc@{}}\toprule 
\multirow{2}{*}{Method} & \multicolumn{6}{c}{Generator} & \phantom{a} & \multicolumn{5}{c}{Render} \\
\cmidrule{2-7} \cmidrule{9-13}
& ExpFD $\downarrow$ & AngleFD $\downarrow$ & TransFD $\downarrow$ & LMD $\downarrow$ & AVOffset $\downarrow$ & AVConf $\uparrow$ && SSIM $\uparrow$ & CPBD $\uparrow$ & PSNR $\uparrow$ & FID $\downarrow$ & CSIM $\uparrow$ \\ 
\midrule
\textcolor{TRed}{Listener} + \textcolor{TBlue}{Speaker} & 13.954 & 6.515 & 5.537 & \textbf{9.325} & \textbf{0.193} & \textbf{2.764} && 0.641 & 0.163 & 17.938 & 41.929 & 0.200 \\
Conversational Agent & \textbf{13.293} & \textbf{5.669} & \textbf{5.434} & 9.430 & 0.362 & 2.418 && 0.652 & 0.164 & 18.448 & 41.842 & 0.196 \\
\bottomrule
\end{tabular}
\end{table*}

\textbf{Conversational Head Generation}\quad To assess the distinctness of the speaker/listener switches agent and our carefully designed agent, we aggregated the best listeners and speakers in ViCo-X dataset to build a baseline and compare with our method. Results in~\cref{tab:agent_resultsx} reveal that the performance of our conversational agent was superior to the baseline on the majority of benchmarks. A plausible explanation for the decline in performance observed in LMD, AVOffset, and AVConf could be attributed to our conversational agent's emphasis on fostering a smoother transition in role alternation. This prioritization aims to enhance naturalness and rationality in conversations.

\subsection{Qualitative Results}
Further, to gain more intuitive insight into these methods, we also visualized the results of the generations to compare the differences between the different configurations.

\begin{figure*}
    \centering
    \includegraphics[width=0.99\linewidth,page=1]{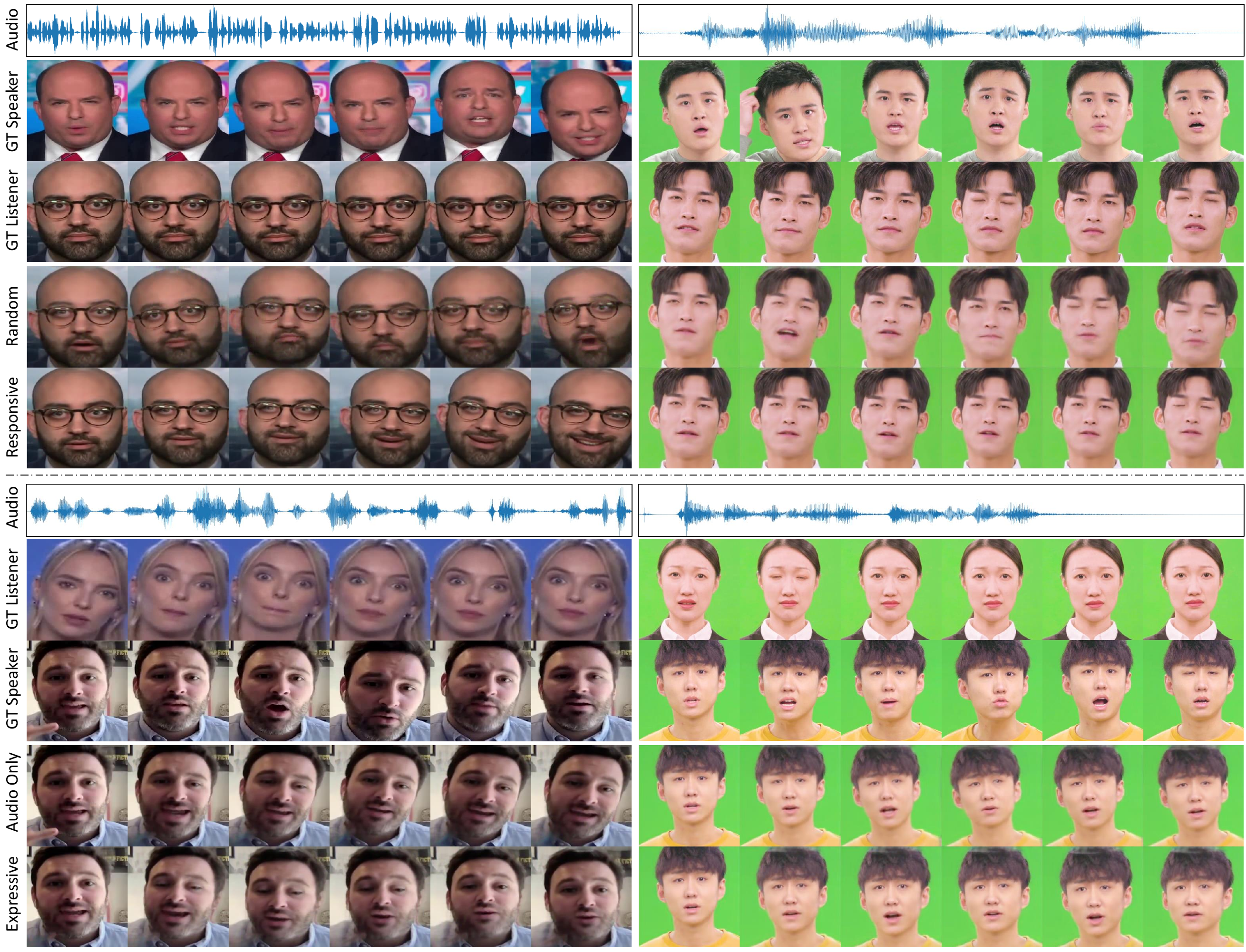}
    \caption{Qualitative results for listening head (upper) and talking head generation (lower) on ViCo (left) and ViCo-X (right). The ``Mirror'' row in the listening head generation section is not shown because it merely utilizes the speaker's motion and expression. The ``Expressive'' row in the talking head generation section denotes the speaker generated with both audio and listener visual signals.}
    \label{fig:qr_lhg_thg}
\end{figure*}

\Cref{fig:qr_lhg_thg} shows the results of listening head generation and talking head generation on ViCo and ViCo-X dataset. For each task and dataset, a random sequence was chosen and then down-sampled to six frames to qualitatively visualize the generated results against the ground truth video. The figure reveals that our model is generally able to capture listener and speaker patterns, such as eye, mouth, and head motion, \etc; even if they may differ from the ground truth but still convey rationality. Overall, in terms of visual plausibility, our results outperform other methods, showcasing more convincing outcomes characterized by realistic head motions and expressive changes.

\begin{figure*}
    \centering
    \includegraphics[width=0.99\linewidth,page=2]{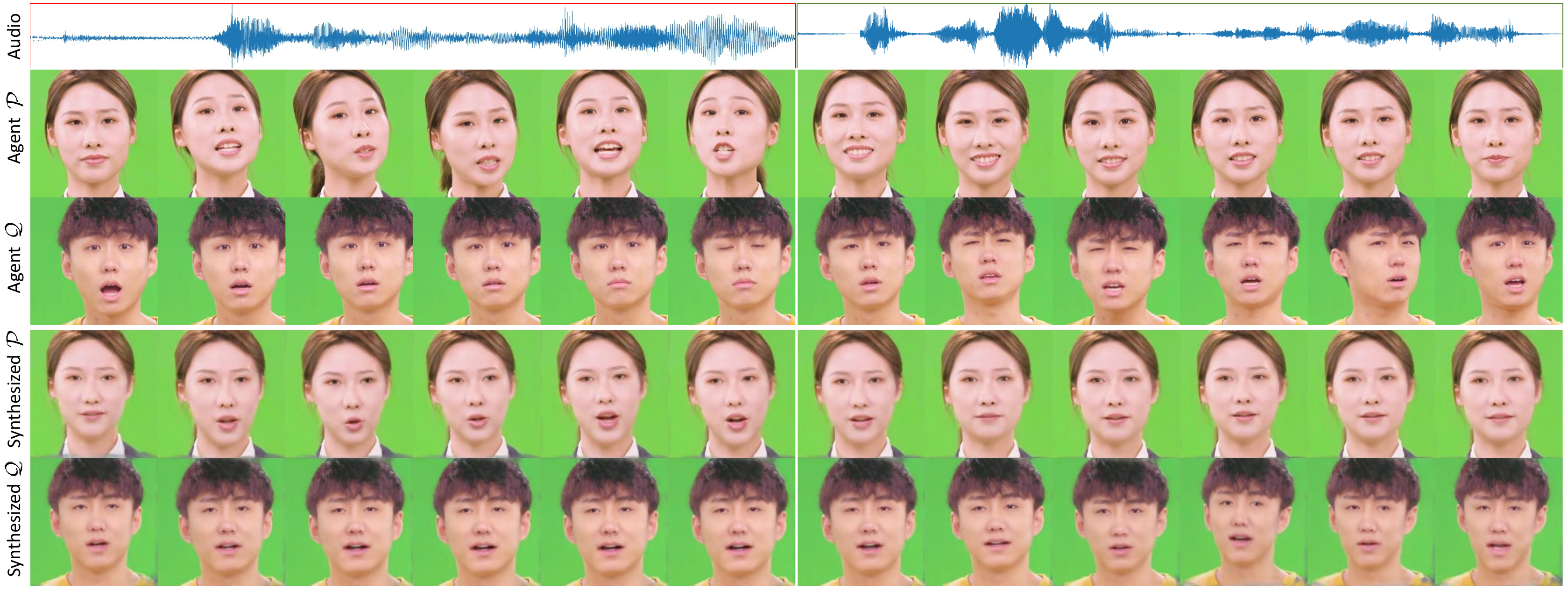}
    \caption{Conversational head generation results on ViCo-X. The audio with a \textcolor{AudioRed}{red border} signifies current speaker is $\cP$, and the \textcolor{AudioGreen}{green border} denotes current speaker is $\cQ$. The last two rows show our synthesized ``Conversational Agent''. When synthesizing one interlocutor, the other's ground-truth visual signals are given.}
    \label{fig:qr_cag}
\end{figure*}

We plot the conversational head generation results on ViCo-X dataset in~\cref{fig:qr_cag}. We employ the same processing strategy as described earlier. Notably, during role alternation, our method exhibits a smoother transition in both motion and expression changes, resulting in a more natural performance overall.

\subsection{User Study}
Given one interlocutor's ground-truth video and the other's synthesized video, twenty volunteers participated in the study and were asked to rate on the total ViCo-X dataset. We adopt the commonly-used five-point scale for each instance: score 1 indicates the synthesized heads cannot speak expressively or listen responsively, while score 5 implies the generated heads are consistent with human subjective perceptions, and even able to communicate with real people. In this test, our synthesized heads get an average score of $3.35\pm 0.12$, which is superior to the baseline \textcolor{TRed}{Listener} + \textcolor{TBlue}{Speaker} $3.02\pm 0.35$. This verifies that our model is capable to mimic the interlocutor behaviors in conversation and results in responsive, vivid, and natural conversational heads. In addition, we also explore the upper bounds of the ViCo-X dataset. The score rated on ground-truth videos is $4.4\pm 0.10$, indicating there is ample room to promote.

\section{Conclusion}
\label{sec:conclusion}
In this paper, we address the problem of synthesizing conversational interaction in current digital humans by proposing two datasets ViCo and ViCo-X. On ViCo, we propose the responsive listening head generation task, which involves generating video clips that respond attentively to a speaker by comprehending their facial signals and voices. Additionally, we introduce the expressive speaker generation task to enable "conversational" speakers to perceive the listener's reactions while speaking. And on ViCo-X, we define the conversational head generation task, which aims to synthesize a conversational agent with the other interlocutor's behaviors. With agent modeling, digital humans can talk, listen, and collaborate with each other to complete the conversation.

We anticipate that ViCo and ViCo-X would prove valuable in human-computer interaction research and virtual human applications. Given the prevalence of communication across various domains such as doctor--patient interactions, teacher--student dialogues, salesperson--customer exchanges, and more, our proposed conversational head generation task fills a crucial void in face-to-face communication modeling. This advancement holds the potential to drive applications in those scenarios, opening new avenues for development and innovation.

\bibliographystyle{IEEEtran}
\bibliography{egbib}
\end{document}